\documentclass[letterpaper, 10 pt, conference]{ieeeconf}  

\usepackage[linesnumbered,ruled,noend,vlined]{algorithm2e}

\SetCommentSty{mycommfont}

\usepackage{times}
\usepackage{adjustbox} 
\usepackage[numbers]{natbib}
\usepackage{multicol}
\usepackage{multirow}
\usepackage[bookmarks=true]{hyperref}
\usepackage{cuted}
\usepackage{lipsum}
\usepackage{comment}
\usepackage{balance}
\usepackage{graphicx}
\usepackage[table]{xcolor}
\usepackage{caption}
\usepackage{listings}
\usepackage{graphicx}
\usepackage[font=small,labelfont=bf]{caption}
\usepackage{subcaption}
\usepackage{xcolor} 
\usepackage{bbding}

\usepackage{pifont}

\newcommand{\redcross}{\textcolor{red}{\ding{55}}}
  \definecolor{darkgray}{RGB}{50, 50, 50}
\definecolor{brightorange}{RGB}{255, 165, 0}
\definecolor{deeporange}{RGB}{255, 140, 0}
\usepackage{xcolor}

\definecolor{codegreen}{rgb}{0,0.6,0}
\definecolor{codegray}{rgb}{0.5,0.5,0.5}
\definecolor{codepurple}{rgb}{0.58,0,0.82}
\definecolor{backcolour}{rgb}{0.95,0.95,0.92}

\lstdefinestyle{mystyle}{
    backgroundcolor=\color{backcolour},
    commentstyle=\color{codegreen},
    keywordstyle=\color{magenta},
    numberstyle=\tiny\color{codegray},
    stringstyle=\color{codepurple},
    basicstyle=\ttfamily,
    breaklines=true,
    captionpos=b,
    keepspaces=true,
    numbers=left,
    numbersep=5pt,
    showstringspaces=false,
    tabsize=2
}
\usepackage{minted}
\usepackage{xcolor}
\definecolor{bg}{HTML}{f2f2ea}
\usepackage{float} 

\usepackage{seqsplit}

\IEEEoverridecommandlockouts                              

\usepackage{tabularx} 
\usepackage{xcolor}

\usepackage{algs}
\usepackage{amsmath} 
\usepackage{amssymb}  
\usepackage{cuted}   
\usepackage{booktabs,caption} 
\usepackage{dblfloatfix} 
\DeclareMathOperator*{\argmin}{arg\,min}

\title{\Large \bf
Robust and Efficient MuJoCo-based Model Predictive Control via \\  Web of Affine Spaces Derivatives
}

\author{Chen Liang$^{1}$ and Daniel Rakita$^{1}$
\thanks{$^{1}$Authors are in the Department of Computer Science, Yale University, New Haven, CT 06520, USA. {\tt\small \{dylan.liang, daniel.rakita\}@yale.edu}. This work was supported by Office of Naval Research award N00014-24-1-2124.}%
}  

\begin{document}

\maketitle
\thispagestyle{empty}
\pagestyle{empty}


\begin{abstract}
MuJoCo is a powerful and efficient physics simulator widely used in robotics. One common way it is applied in practice is through Model Predictive Control (MPC), which uses repeated rollouts of the simulator to optimize future actions and generate responsive control policies in real time.  To make this process more accessible, the open source library MuJoCo MPC (MJPC) provides ready-to-use MPC algorithms and implementations built directly on top of the MuJoCo simulator.  However, MJPC relies on finite differencing (FD) to compute derivatives through the underlying MuJoCo simulator, which is often a key bottleneck that can make it prohibitively costly for time-sensitive tasks, especially in high-DOF systems or complex scenes.  In this paper, we introduce the use of Web of Affine Spaces (WASP) derivatives within MJPC as a drop-in replacement for FD. WASP is a recently developed approach for efficiently computing sequences of accurate derivative approximations. By reusing information from prior, related derivative calculations, WASP accelerates and stabilizes the computation of new derivatives, making it especially well suited for MPC's iterative, fine-grained updates over time. We evaluate WASP across a diverse suite of MJPC tasks spanning multiple robot embodiments. Our results suggest that WASP derivatives are particularly effective in MJPC: it integrates seamlessly across tasks, delivers consistently robust performance, and achieves up to a 2$\mathsf{x}$ speedup compared to an FD backend when used with derivative-based planners, such as iLQG. In addition, WASP-based MPC outperforms MJPC's stochastic sampling-based planners on our evaluation tasks, offering both greater efficiency and reliability. To support adoption and future research, we release an open-source implementation of MJPC with WASP derivatives fully integrated.\footnote{\href{https://github.com/chen-dylan-liang/mujoco_wasp_mpc}{https://github.com/chen-dylan-liang/mujoco\_wasp\_mpc}} \footnote{\href{https://github.com/chen-dylan-liang/wasp-differentiated-mujoco}{https://github.com/chen-dylan-liang/wasp-differentiated-mujoco}} 
\end{abstract}


\section{Introduction}
\label{sec:introduction}

MuJoCo is a powerful and efficient physics simulator widely used in robotics~\cite{mujoco}. One way it is applied in practice is through Model Predictive Control (MPC), which uses repeated rollouts of the simulator to optimize future actions and generate responsive control policies in real time. MPC can be implemented on top of a simulator like MuJoCo by differentiating through the simulator, allowing the outer optimization loop to follow derivative information and continuously ``surf'' the function landscape downhill toward a local minimum over time. To make this process more accessible, the open source library MuJoCo MPC (MJPC) provides ready-to-use MPC algorithms and implementations built directly on top of the MuJoCo simulator~\cite{mjpc}.    

In theory, it is possible to leverage automatic differentiation (AD) tools to conveniently compute exact derivatives of the underlying simulator.  For instance, MuJoCo offers XLA and Warp backends to use state-of-the-art AD implementations.  However, in the context of MPC, these exact derivatives often prove too narrow in scope, as they capture only highly local sensitivities of the nonlinear dynamics. Because MPC repeatedly differentiates through short-horizon rollouts of these complex dynamics, the resulting derivatives can become excessively sharp or ill-conditioned, which frequently leads to numerical instability.

In turn, MJPC relies solely on finite differencing (FD) to compute derivatives through the underlying MuJoCo simulator.  FD estimates derivatives by perturbing each input dimension and measuring the resulting change in the simulator’s output.  At a high level, FD seems well suited for MPC: it is straightforward to apply, requiring only standard forward passes without modifications to the simulator’s source code, and its small approximation error in the derivatives can contribute to greater numerical stability by smoothing out sharp or ill-conditioned derivatives that arise from exact differentiation.  

\begin{figure}[t]
    \centering
    \begin{subfigure}[b]{0.32\columnwidth}
        \centering
         \caption{Quadrotor}
        \includegraphics[width=\textwidth]{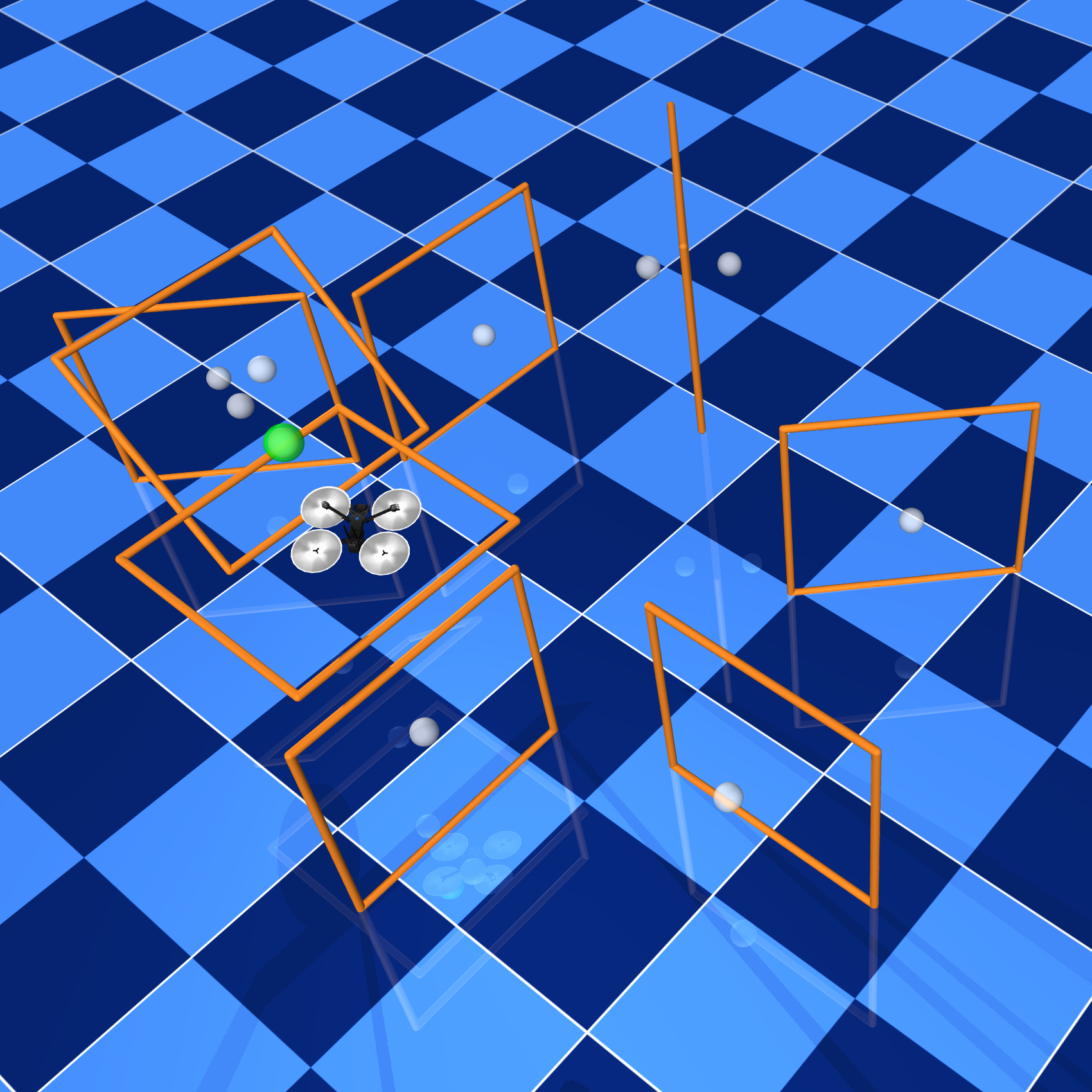}
        \label{fig:sub1}
    \end{subfigure}
  \hfill
    \begin{subfigure}[b]{0.32\columnwidth}
        \centering
         \caption{Swimmer}
        \includegraphics[width=\textwidth]{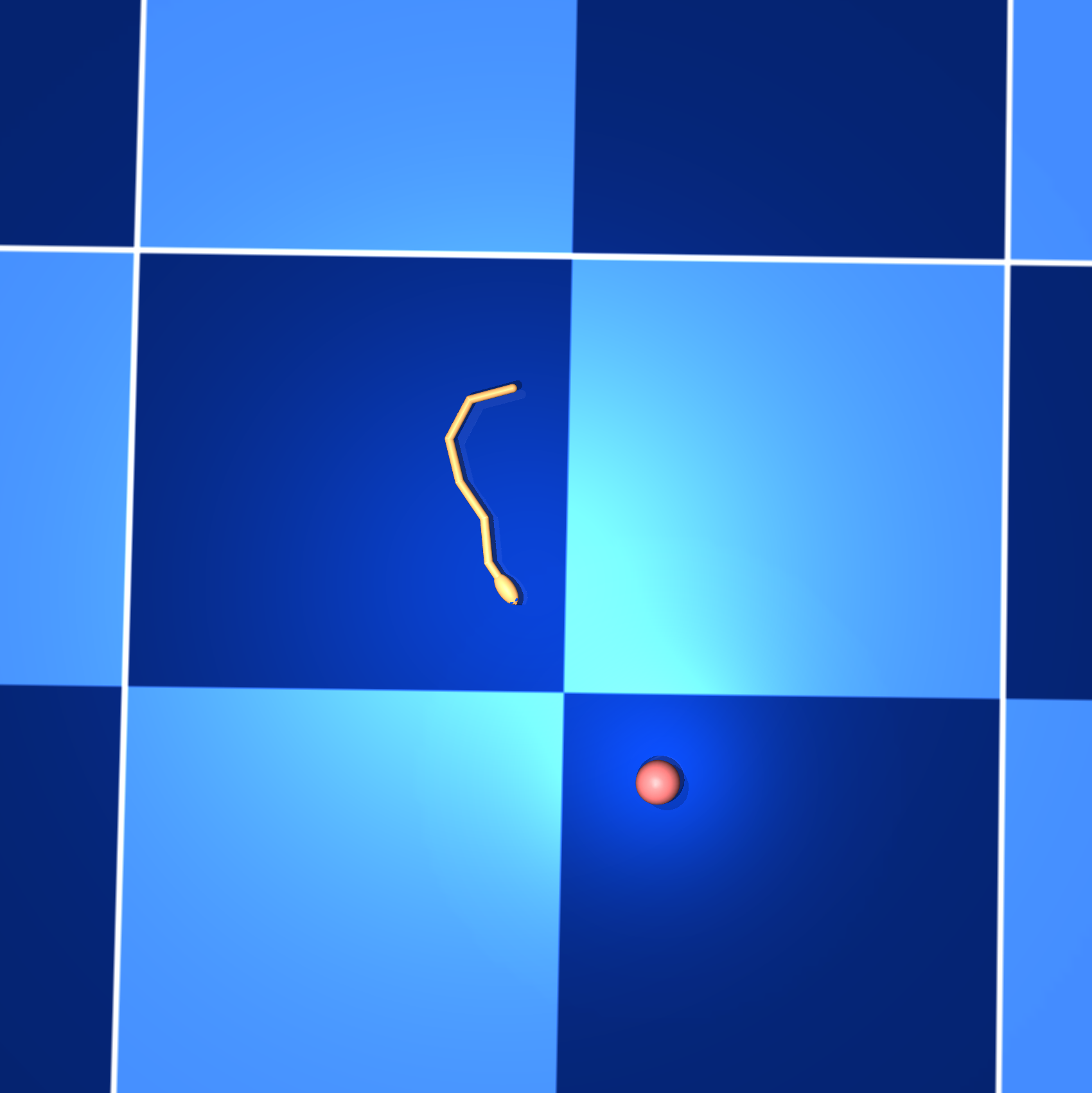}
        \label{fig:sub2}
    \end{subfigure}
    \hfill
    \begin{subfigure}[b]{0.32\columnwidth}
        \centering
        \caption{Quadruped Climb}
        \includegraphics[width=\textwidth]{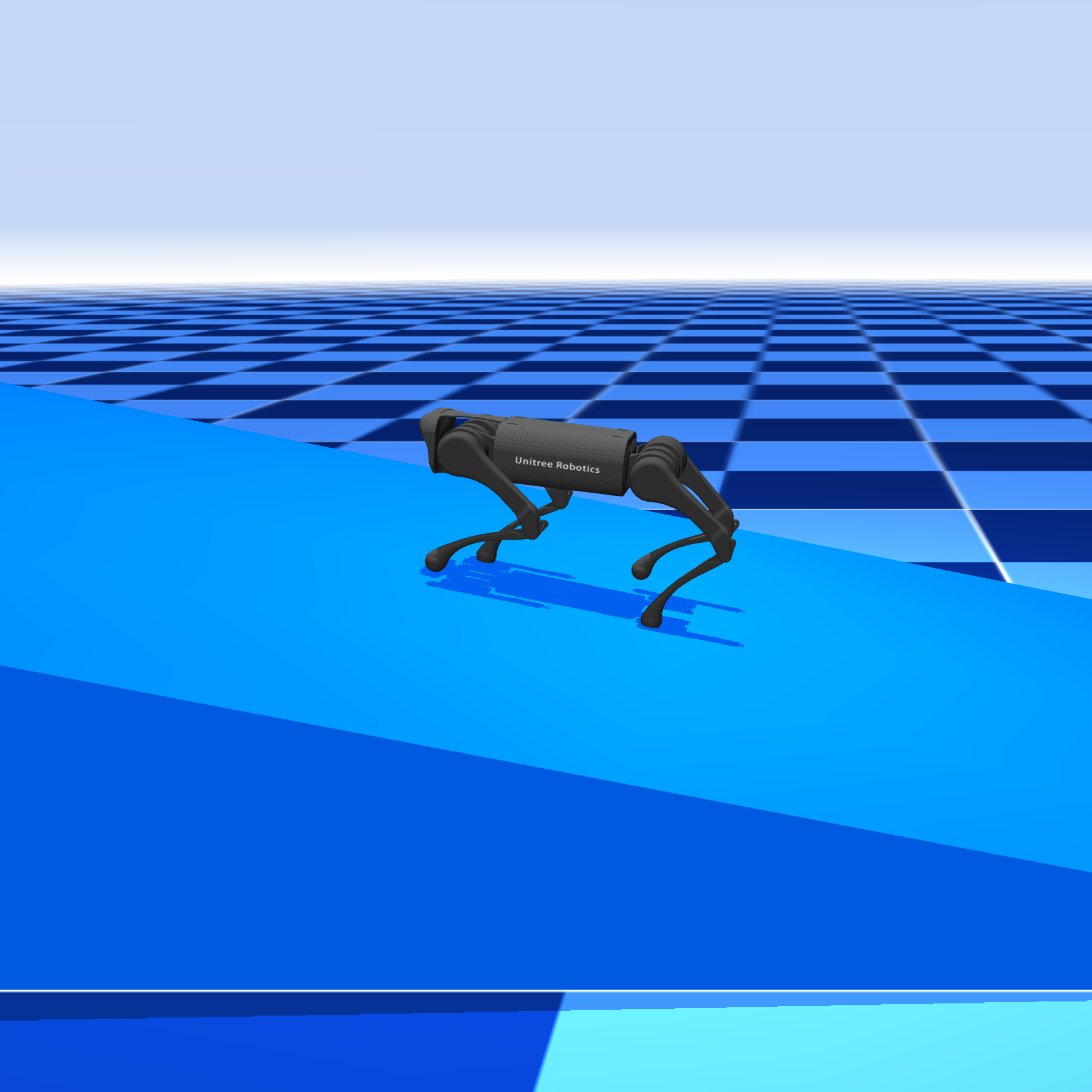}
        \label{fig:sub3}
    \end{subfigure}
    
    \vspace{-12pt}  
    \begin{subfigure}[b]{0.32\columnwidth}
        \centering
        \caption{Quadruped Move}
        \includegraphics[width=\textwidth]{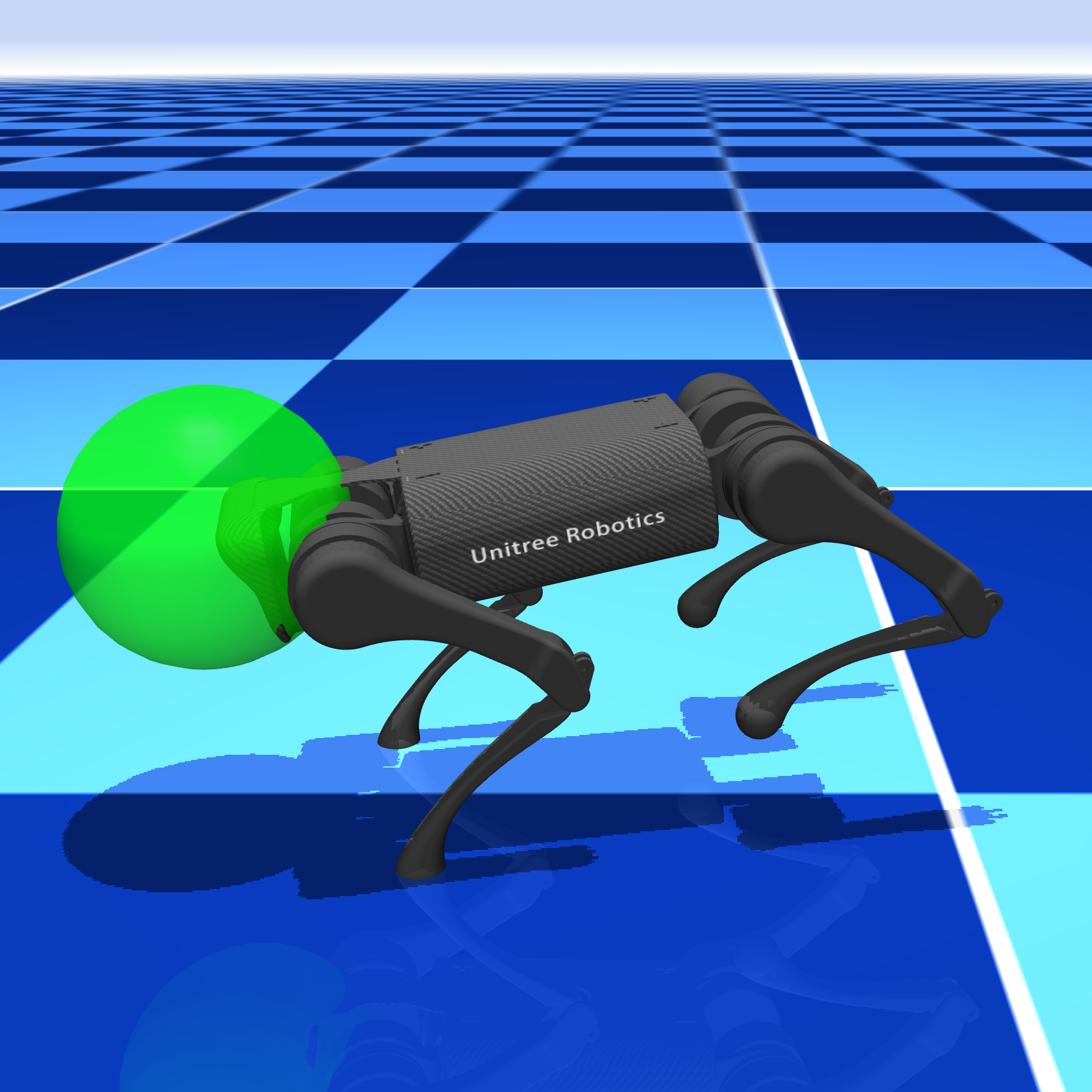}
        \label{fig:sub5}
    \end{subfigure}
  \hfill
  \begin{subfigure}[b]{0.32\columnwidth}
        \centering
         \caption{Biped Balance}
        \includegraphics[width=\textwidth]{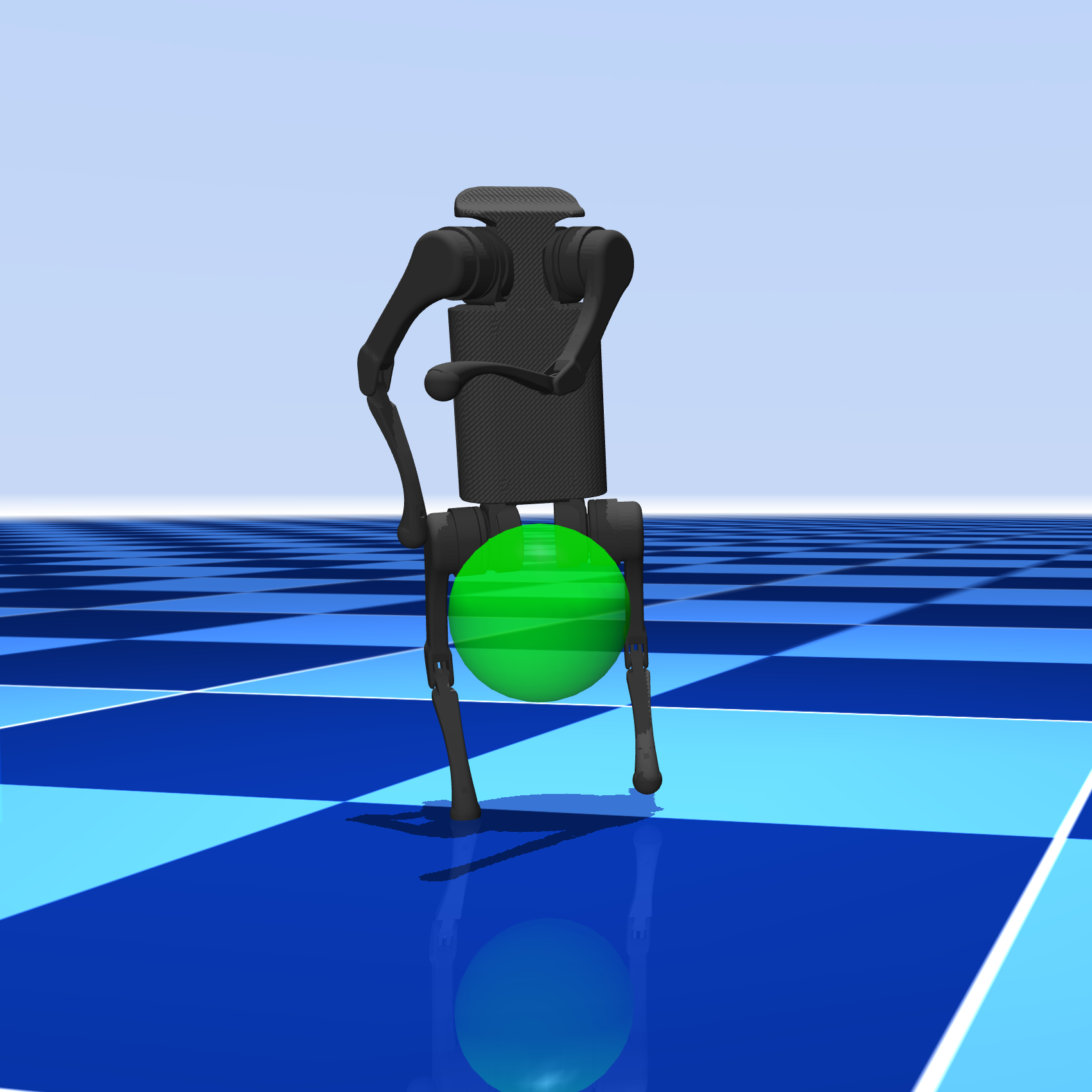}
        \label{fig:sub4}
    \end{subfigure}
\hfill
    \begin{subfigure}[b]{0.32\columnwidth}
        \centering
             \caption{Humanoid Walk}
        \includegraphics[width=\textwidth]{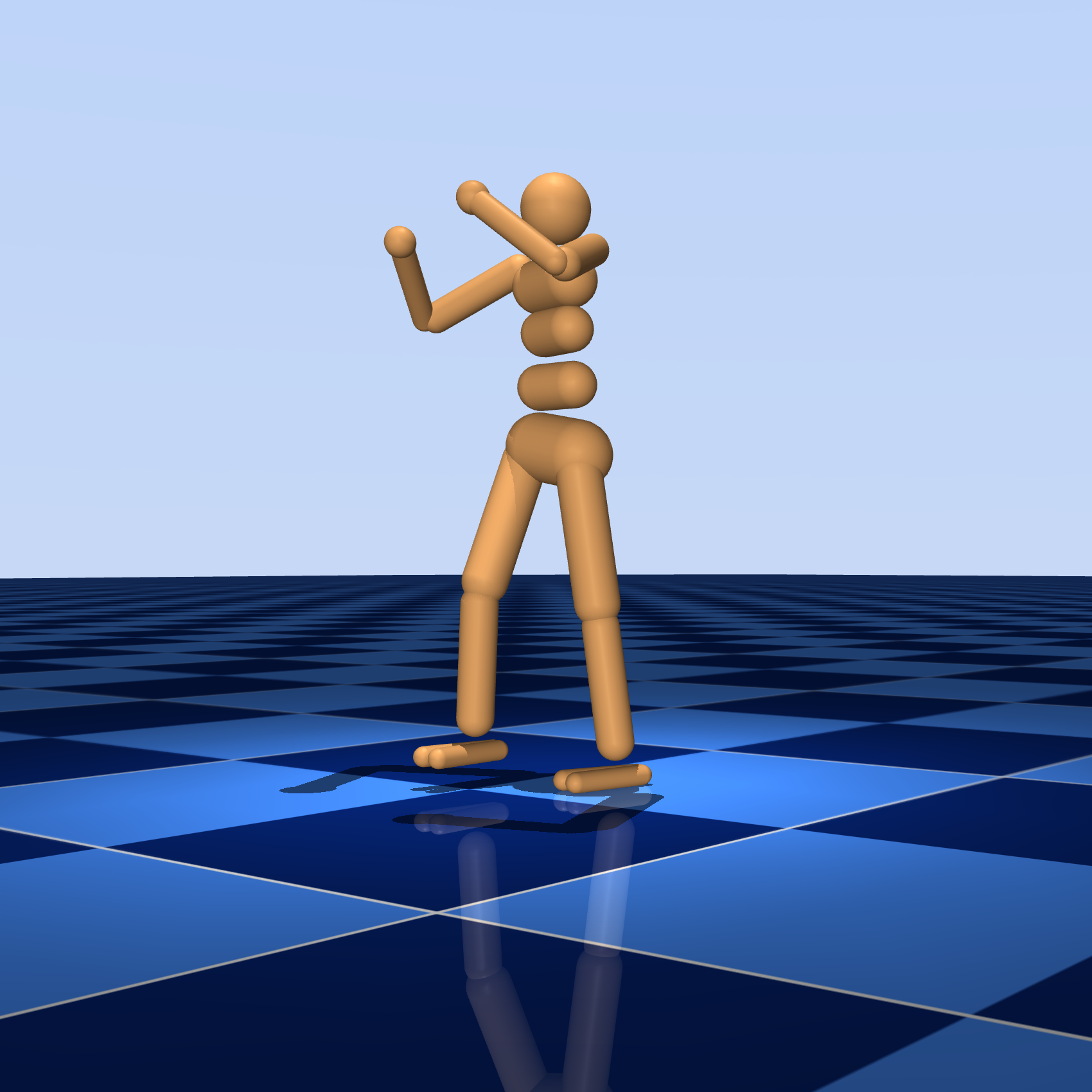}
        \label{fig:sub6}
    \end{subfigure}
    \vspace{-10pt}  
    \caption{Visuals of $10$ tasks evaluated to demonstrate the efficacy of WASP-based derivatives in MJPC. Relative to finite differences, WASP accelerates the total compute time by up to 2$\mathsf{x}$ while maintaining robust performance. }
    \label{fig:teaser}
    \vspace{-0.6cm}
\end{figure}

Despite these advantages, FD becomes inefficient for high-DOF systems or complex scenes. Each input dimension must be perturbed independently, so the number of simulator calls grows linearly with the dimensionality of the state and action spaces. For robots with many joints or environments featuring rich contact dynamics, this approach can require hundreds or even thousands of additional rollouts for a single derivative evaluation. Although parallelization can mitigate part of this cost, the unfavorable scaling remains a fundamental limitation. As a result, the computational burden often dominates the control loop, making it challenging for MJPC to sustain the fast update rates required for real-time robotics.

In this paper, we introduce the use of Web of Affine Spaces (WASP) derivatives within MJPC as a drop-in replacement for FD. WASP is a recently developed method for efficiently computing sequences of approximate derivatives by reusing information from prior, related evaluations, enabling faster computation and improved numerical stability~\cite{wasp}. Our premise is that these properties make WASP a good fit for MPC, where derivatives are recomputed repeatedly along closely related trajectories. However, prior evaluations of WASP have been limited to kinematics-based functions, and its effectiveness in dynamics-based MPC settings remains untested. A key contribution of this work is to evaluate the extent to which WASP’s advantages carry over to MPC problems involving full physics simulation.



We evaluate WASP across a diverse suite of MJPC tasks spanning multiple robot embodiments (some shown in Figure~\ref{fig:teaser}). Our results suggest that WASP derivatives are particularly effective in MJPC: it integrates seamlessly across tasks, delivers consistently robust performance, and achieves up to a 2$\mathsf{x}$ speedup compared to an FD backend when used with derivative-based planners, such as iLQG. In addition, WASP-based MPC outperforms MJPC's stochastic sampling-based planners on our evaluation tasks, offering both greater efficiency and reliability. 

To support adoption and further research, we provide an open-source implementation of MJPC that incorporates WASP derivatives. This release makes it easy for practitioners to experiment with WASP as a drop-in replacement for FD, enabling immediate evaluation of its efficiency, stability, and accuracy tradeoffs in real-world control tasks.  

\section{Background}
\label{sec:related_works}
In this section, we present the background relevant to our work. We begin with a primer on Model Predictive Control, outlining its optimization formulation and common solution strategies. We then review alternatives to finite differences (FD) for computing model derivatives, with a focus on the strengths and limitations of Automatic Differentiation (AD). Finally, we formalize the problem addressed in this paper.

\subsection{Model Predictive Control Primer}
\label{sec:mpc_primer}

Model Predictive Control (MPC) is a control strategy that repeatedly simulates forward dynamics to predict future states, then optimizes a short horizon of actions to minimize a cost function before executing the first action and replanning.  Specifically, MPC typically reasons over the following components: 


\begin{itemize}
    \item A fixed time-step, $\Delta t$.
    \item An ordered index set, $\mathcal{I} \equiv \{0,1,2,\dots,T\}$, where each index corresponds to a physical time point in the discrete horizon $\{0, \Delta t, 2\Delta t, \dots, T\Delta t\}$.
    \item An ordered set of states, $X \equiv \{ \mathbf{x}_i \mid i \in \mathcal{I} \}$.
    \item An ordered set of controls, $U \equiv \{ \mathbf{u}_i \mid i \in \mathcal{I} \setminus \{T\} \}$. The exclusion of $T$ reflects that no control is applied at the terminal state.
    \item A system dynamics function, $f$, that maps a state and action pair at one index to the state at the next index, i.e., $f(\mathbf{x}_i, \mathbf{u}_i) = \mathbf{x}_{i+1}$.
    \item A scalar cost function, $J$, defined as a sum of stage and terminal costs:
    
    $$
    J(X,U) = \sum_{i=0}^{T-1} \underbrace{\ell(\mathbf{x}_i,\mathbf{u}_i)}_{\text{stage cost}} + \underbrace{\ell_T(\mathbf{x}_T)}_{\text{terminal cost}},
    $$
    
      where $\ell:\mathbb{R}^n\times\mathbb{R}^m\to\mathbb{R}$  and $\ell_T:\mathbb{R}^n\to\mathbb{R}$.
\end{itemize}

The goal of MPC is to compute controls $U$ that, through the system dynamics $f$, produce corresponding states $X$ minimize (or at least reduce) the cost $J$. This optimization is repeated in real time: the first control $\mathbf{u}_0$ from the current solution is applied to the robot, and the search for the next solution begins immediately thereafter.

\subsection{Strategies for Solving MPC}
\label{sec:mpc_strategies}

At a high level, methods for solving the MPC problem specified in \S\ref{sec:mpc_primer} are grouped into two categories: (1) derivative-based methods and (2) stochastic sampling-based methods.  Derivative-based methods typically involve the following steps:

\begin{enumerate}
    \item Rollout candidate states $X$ using candidate controls $U$ and the system dynamics function $f$.  Typically, $U$ is taken from the most recent solution as a warm start.
    \item Compute $\frac{\partial f}{\partial \mathbf{x}_i} \in \mathbb{R}^{d_x \times d_x}$ for all $i \in \mathcal{I}$ and $\frac{\partial f}{\partial \mathbf{u}_i} \in \mathbb{R}^{d_x \times d_u}$ for all $i \in \mathcal{I} \setminus \{T\}$.  These are called \textit{model derivatives}. For some algorithms, second-order model derivatives are also computed or approximated at this step.
    \item Compute $\frac{\partial \ell}{\partial \mathbf{x}_i} \in \mathbb{R}^{1 \times d_x}$ and $\frac{\partial \ell}{\partial \mathbf{u}_i} \in \mathbb{R}^{1 \times d_u}$.  These are called \textit{cost derivatives}.  For some algorithms, second-order cost derivatives are also computed at this step.
    \item Do backward pass using information from Steps 2 and 3 to compute $\frac{\partial J}{\partial \mathbf{u}_T}, \frac{\partial J}{\partial \mathbf{u}_{T-1}}, ..., \frac{\partial J}{\partial \mathbf{u}_0}$.
    \item Take some step over all controls, e.g., $\mathbf{u}_i \leftarrow \mathbf{u}_i + \alpha \frac{\partial J}{\partial \mathbf{u}_i}^\top$.  The actual step here is algorithm-dependent and can use various different optimization strategies (e.g., line search or trust region method).
    \item Optionally, re-rollout candidate states $X$ based on the just updated controls $U$ using the system dynamics function $f$ and repeat Steps 2--6 until some termination condition is reached.
    \item Return answer, $U^*$ and send the first command, $\mathbf{u}_0^*$ to the robot.
\end{enumerate}

In contrast, stochastic sampling-based methods, optimize by evaluating and refining a distribution over candidate control sequences rather than relying on derivatives. These methods typically involve the following steps:

\begin{enumerate}
    \item Initialize a distribution over candidate control sequences $U$, often centered around the most recent solution for warm starting.
    \item Sample a batch of candidate control sequences $\{U^{(k)}\}_{k=1}^K$ from the distribution.
    \item For each sampled sequence $U^{(k)}$, rollout the corresponding states $X^{(k)}$ using the system dynamics function $f$.
    \item Evaluate the cost $J(X^{(k)}, U^{(k)})$ for each trajectory.
    \item Update the sampling distribution parameters based on the best-performing trajectories.
    \item Repeat Steps 2--5 until some termination condition is met (e.g., maximum iterations or convergence of the distribution).
    \item Return the best control sequence $U^*$ found and send the first command, $\mathbf{u}_0^*$, to the robot.
\end{enumerate}

Both derivative-based and stochastic sampling-based strategies have proven effective, but they commonly excel in different regimes. For instance, derivative-based methods achieve fast convergence when accurate model derivatives are available and the function landscape is reasonably smooth, while stochastic approaches may offer more robustness in highly nonlinear or discontinuous settings where derivatives are unreliable. 


In this paper, our focus is on improving derivative-based methods for MPC, though we also compare against stochastic-based methods in our evaluation.


\subsection{MuJuCo MPC Algorithms}

In MJPC, the system dynamics function $f$ is provided by the MuJoCo physics engine itself. MJPC includes implementations of both derivative-based and stochastic sampling-based methods, offering a spectrum of planners with different trade-offs in speed, robustness, and sample efficiency.

Specifically, The derivative-based planners include (1) gradient descent~\cite{gdplanner}; and (2) iterative Linear Quadratic Gaussian (iLQG)~\cite{ilqg}. The stochastic sampling-based planners include (1) predictive sampling~\cite{mjpc}; (2) robust sampling~\cite{mppi}; (3) cross-entropy~\cite{cem}; and (4) sample gradient methods~\cite{zoopt}.  Our evaluation in this paper (\S\ref{sec:evaluation}) systematically assesses all of these planners provided in MJPC.

\subsection{MuJoCo MPC Differentiation}

Our work is primarily addressing the potential bottleneck in Step 2 of the derivative-based strategy steps in \S\ref{sec:mpc_strategies}.  While Step 3 also involves computing derivatives, these cost derivatives are typically computed analytically and, thus, do not require further investigation.

MuJoCo currently computes model derivatives using \textit{finite differencing} (FD).  This process generally takes the following form:

\begin{equation}
    \frac{\partial f}{\partial \mathbf{x}_i}[:,j] \approx \frac{f(\mathbf{x}_i + \epsilon\mathbf{e}_j, \mathbf{u}_i) - f(\mathbf{x}_i, \mathbf{u}_i)}{\epsilon}
    \label{eq:fd_state}
\end{equation}    

\begin{equation}
    \frac{\partial f}{\partial \mathbf{u}_i}[:,k] \approx \frac{f(\mathbf{x}_i, \mathbf{u}_i + \epsilon\mathbf{e}_k) - f(\mathbf{x}_i, \mathbf{u}_i)}{\epsilon},
    \label{eq:fd_control}
\end{equation}

\noindent where $\frac{\partial f}{\partial \mathbf{x}_i}[:,j]$ and $\frac{\partial f}{\partial \mathbf{u}_i}[:,k]$ refer to the $j$-th and $k$-th columns in these matrices, respectively, $\mathbf{e}_j$ and $\mathbf{e}_k$ denote one-hot vectors where only the $j$-th and $k$-th elements are one and all others are zero, and $\epsilon$ is a small scalar value (e.g., $1e^{-5}$).

Note that the FD processes in Equations~\ref{eq:fd_state} and \ref{eq:fd_control} must be run for all $d_x$ columns in $\frac{\partial f}{\partial \mathbf{x}_i}$, all $d_u$ columns in $\frac{\partial f}{\partial \mathbf{u}_i}$, and all $T+1$ time points.  Thus, a single update to $U$ requires $d_x*(T+1) + d_u*T+1$ calls to the system dynamics function $f$ which, itself, can be fairly expensive.  

While parallelization may help in practice, it typically applies to only one loop dimension. In MJPC, for example, computations are distributed across CPU threads along the time index $i$. Even so, the workload may remain prohibitive when the state or control dimension ($n$ or $m$) is large or when evaluating the dynamics $f$ is particularly costly.

\subsection{Other Derivative Computation Strategies}
Automatic Differentiation (AD) is a natural alternative to FD, supported by mature software implementations~\cite{autodiff}. AD can be applied in either forward mode or reverse mode~\cite{ad_tutorial}, and both have gained popularity in robotics. Several differentiable simulators, such as gradSim~\cite{gradsim}, PlasticineLab~\cite{softbodysim}, and \texttt{lcp-physics}~\cite{endtoenddiff}, employ reverse-mode AD to compute model derivatives. More recently, \texttt{ad-trait}~\cite{adtrait} introduced the first operator-overloading AD library in Rust, supporting both modes and integrating with a Rust-based robotics library.  MuJoCo offers XLA and Warp backends to use state-of-the-art AD implementations, though these have not been integrated into MJPC. 

Despite these advances, \citet{ad_drawbacks} highlight that characteristics of contact dynamics, such as stiffness and discontinuities, can compromise the utility of AD, producing non-informative derivatives and numerical instability. Exact derivatives often prove too narrow in scope, as they capture only highly local sensitivities of the nonlinear dynamics~\cite{callejo2014performance}. Because MPC repeatedly differentiates through short-horizon rollouts of these complex dynamics, the resulting derivatives can become excessively sharp or ill-conditioned, which frequently leads to numerical instability. 

Building on MuJoCo's FD implementation, Zhang et al.~\cite{real_world_mpc} proposed a heuristic to accelerate computation by skipping derivative evaluations at selected time points and interpolating them from nearby evaluations along the planning horizon. This strategy is complementary to our approach: their method reduces costs from long horizons, while ours targets the scaling issues introduced by high DoF systems. In practice, the two techniques could be combined, and we plan to explore these connections in future work. 

\subsection{Problem Statement} 

The primary bottleneck in derivative-based MPC arises in Step~2 of the solution process outlined in \S\ref{sec:mpc_strategies}, where model derivatives must be repeatedly computed through the system dynamics function. Our goal is to accelerate these derivative computations in MJPC, thereby reducing the overall time required for each MPC iteration. Our goal is to continue to treat the MuJoCo physics engine as a black-box model, without requiring any modifications to its source code. Crucially, any such acceleration must preserve robustness: the resulting control sequences should continue to yield stable, low-cost trajectories, ensuring that long-horizon task performance, as measured by returns from $J$, remains uncompromised.

\section{Web of Affine Spaces Derivatives}
\label{sec:wasp-intro}

Our strategy for accelerating model derivatives while maintaining robustness in MJPC is to incorporate the use of Web of Affine Spaces (WASP) derivatives.  In this section, we overview the WASP approach.  

\subsection{Coherence-based Approximate Derivatives}
\label{sec:coherence}

At a high level, WASP is an approach designed to compute a sequence of approximate derivatives:

\vspace{-3pt}
\begin{equation}
\dots, \ \hat{\frac{\partial f}{\partial \mathbf{x}}}\bigg|_{\mathbf{x}_{k-1}}, \  \hat{\frac{\partial f}{\partial \mathbf{x}}}\bigg|_{\mathbf{x}_{k}}, \ \hat{\frac{\partial f}{\partial \mathbf{x}}}\bigg|_{\mathbf{x}_{k+1}}, \ \dots \ 
\end{equation}

\noindent for some differentiable function $f$ and sequence of inputs $( \ \cdots \mathbf{x}_{k-1}, \mathbf{x}_{k}, \mathbf{x}_{k+1}, \cdots \ )$.  

The key insight associated with WASP is that derivatives need not be computed from scratch. In many applications, especially those rooted in iterative optimization, the function inputs evolve gradually over time, one leading into the next. As a result, the corresponding sequence of derivatives tends to exhibit \textit{temporal or spatial coherence}, i.e., adjacent elements in the sequence share similarities. Thus, by reusing information from previous, closely related derivative computations, WASP seeks to accelerate derivative estimation while preserving accuracy. 

\subsection{WASP Overview}

At its core, the WASP approach formulates derivative estimation as a constrained least-squares problem. Each iteration of the algorithm requires only a single Jacobian-vector product (JVP), which defines an affine subspace guaranteed to contain the true derivative. The optimization then seeks the transpose of an approximate derivative that lies within this affine subspace, enforced via a hard constraint, while simultaneously aligning with prior, related computations encoded in the objective function. 

This optimization is formalized as follows:

\vspace{-5pt}
\begin{equation}
\label{eq:wasp}
\begin{gathered}
\frac{\partial f}{\partial \mathbf{x}}\bigg|_{\mathbf{x}_k}^\top \approx \argmin_{\mathbf{D}^\top} || \Delta\mathbf{X}^\top \mathbf{D}^\top - \hat{\Delta\mathbf{F}}^\top ||^2_F, \\ 
\text{s.t.} \quad \Delta \mathbf{x}_i^\top \mathbf{D}^\top = \Delta \mathbf{f}_i^\top \ .
\end{gathered}
\end{equation}

\noindent Here, $\Delta \mathbf{X} \in \mathbb{R}^{n \times n}$ is referred to as the \textit{tangent matrix}, whose columns represent input directions used to define the local linear neighborhood. The variable $\mathbf{D} \in \mathbb{R}^{m \times n}$ denotes the candidate Jacobian being optimized. The matrix $\hat{\Delta \mathbf{F}} \in \mathbb{R}^{m \times n}$ provides approximate JVPs corresponding to the directions in $\Delta \mathbf{X}$, defined as $\hat{\Delta \mathbf{F}} \approx \frac{\partial f}{\partial \mathbf{x}}\big|_{\mathbf{x}_k} \ \Delta \mathbf{X}$.  The hard constraint enforces consistency in one chosen direction, where $\Delta \mathbf{x}_i$ is the $i$-th column of $\Delta \mathbf{X}$, and $\Delta \mathbf{f}_i$ is the corresponding ground-truth JVP at $\mathbf{x}_k$. This single JVP is estimated using finite differencing:

\begin{equation}
\label{eq:finite_diff}
\Delta \mathbf{f}_i = \frac{\partial f}{\partial \mathbf{x}}\bigg|_{\mathbf{x}_k} \ \Delta \mathbf{x}_i \approx \frac{f(\mathbf{x}_k + \epsilon\Delta \mathbf{x}_i) - f(\mathbf{x}_k)}{\epsilon} \ .
\end{equation}

\noindent The goal is for $\hat{\Delta \mathbf{F}}$ to require only a small number of new Jacobian-vector products (JVPs) per derivative computation, while the remaining columns preserve sufficient information from previous evaluations to yield a highly accurate approximation.  

In previous work, \citet{wasp} derive a closed form solution for Equation~\ref{eq:wasp} using a KKT system.  They also show how to precompute and cache certain matrices in this solution to accelerate the solution at runtime.       

\subsection{Control over WASP Approximation}
\label{sec:parameters}
The WASP approach affords a convenient tradeoff between accuracy and efficiency.  At a high level, this tradeoff is dictated by how many JVPs are computed (via function calls to the function $f$) to update $\hat{\Delta \mathbf{F}}$ per approximate solution, $\hat{\frac{\partial f}{\partial \mathbf{x}}}\big|_{\mathbf{x}_k}$.  Using more JVPs yields a more accurate approximation, but at the cost of increased runtime due to more evaluations of $f$.

Four parameters are used to control the number of JVPs used per solution, denoted as $p_{max}, p_{min}, p_\theta, p_n$.  Here, $p_{max}$ and $p_{min}$ set a maximum and minimum on the number of JVPs, respectively.  The $p_\theta$ and $p_n$ stop the JVP calculations and return a result when the current approximate derivative elicits approximate JVPs that sufficiently match the angle and norm of ground truth JVPs, respectively (fully detailed by \citet{wasp}).  Lower values of $p_\theta$ and $p_n$ correspond to stricter accuracy thresholds, leading to more JVPs and higher computational cost.

\section{Implementation Details}
\label{sec:implementation_details}

In this section, we describe how the WASP derivative approach is integrated into the MuJoCo MPC (MJPC) framework. We outline the practical aspects of adapting WASP to MJPC's planning pipeline, highlight tunable parameters that allow users to balance speed and accuracy, and discuss extensions to the MJPC graphical interface that make these capabilities accessible in practice.

\subsection{WASP in MJPC}

In MJPC, WASP derivatives are used to compute the model derivative matrices 
$\frac{\partial f}{\partial \mathbf{x}_i} \in \mathbb{R}^{d_x \times d_x}$ and 
$\frac{\partial f}{\partial \mathbf{u}_i} \in \mathbb{R}^{d_x \times d_u}$ at each time index $i$, 
overviewed in \S\ref{sec:mpc_strategies}. Each time point along the planning horizon is treated as an independent 
instance of WASP. These instances can run independently in parallel and can optionally plug directly into MuJoCo's existing parallelization over the time horizon. This design ensures that the benefits of WASP scale naturally with MJPC's parallel execution model.

Each WASP instance requires its own approximate JVP matrix, $\hat{\Delta \mathbf{F}}$. For model derivatives with respect to states, we maintain 
$\hat{\Delta \mathbf{F}}_{x, i} \in \mathbb{R}^{d_x \times d_x}$, and for derivatives with respect to controls, 
$\hat{\Delta \mathbf{F}}_{u, i} \in \mathbb{R}^{d_x \times d_u}$. These matrices are updated incrementally as new JVPs are computed during planning.

In addition, each WASP instance requires a tangent matrix $\Delta \mathbf{X}$ to define the local linear neighborhood. 
These are $\Delta \mathbf{X}_{x, i} \in \mathbb{R}^{d_x \times d_x}$ for state derivatives and 
$\Delta \mathbf{X}_{u, i} \in \mathbb{R}^{d_u \times d_u}$ for control derivatives. However, since the tangent matrices 
are constant throughout runtime, we provide an option to reuse the same matrices across all time points, reducing 
memory requirements. In this case, only two matrices are maintained globally: 
$\Delta \mathbf{X}_{x} \in \mathbb{R}^{d_x \times d_x}$ and 
$\Delta \mathbf{X}_{u} \in \mathbb{R}^{d_u \times d_u}$. This shared-tangent setting is the default used in our 
evaluation. Following the analysis and recommendations of \citet{wasp}, all tangent matrices are chosen to be 
orthonormal to ensure favorable numerical properties.

Finally, we directly incorporate these changes into the MuJoCo source code alongside the existing FD implementation in 
the C programming language. This integration allows WASP to function as a true drop-in replacement for all downstream 
MuJoCo-based applications, including our present study of MPC in MJPC.

\subsection{Tunable Parameters}

As introduced in \S\ref{sec:parameters}, WASP has four low-level parameters ($p_{max}, p_{min}, p_{\theta}, p_n$) that govern how many JVPs are used per derivative solution, trading off between accuracy and efficiency.  To simplify this interface in MJPC, we expose only two parameters for users to adjust: a \textit{fraction parameter}, denoted as \texttt{frac}, and a \textit{tolerance parameter}, denoted as \texttt{tol}.

Formally, these parameters are defined as follows:

$$
\texttt{frac} = \frac{p_{min}}{p_{max}}, \ \ \ \texttt{tol} = p_{\theta} = p_{n}.
$$

\noindent In other words, \texttt{frac} is a normalized value in the range $(0,1]$ relating $p_{min}$ and $p_{max}$, and $\texttt{tol}$ is a stand-in label when $p_{\theta}$ and $p_{n}$ are equal, as is always the convention in our implementation.  Also, we always fix $p_{max}$ to be the full number of inputs to the function (\citet{wasp} show that WASP reduces to finite differencing at this number of JVPs); thus, \texttt{frac} is only a function of $p_{min}$, making it easier to adjust with respect to the fixed $p_{max}$ on a normalized scale. 

Both parameters can be applied separately to the state and control derivative computations, which we denote as $(\texttt{frac}_x, \texttt{tol}_x)$ and $(\texttt{frac}_u, \texttt{tol}_u)$, respectively.   

\begin{figure}[t!]
    \centering
    \includegraphics[width=0.8\columnwidth]{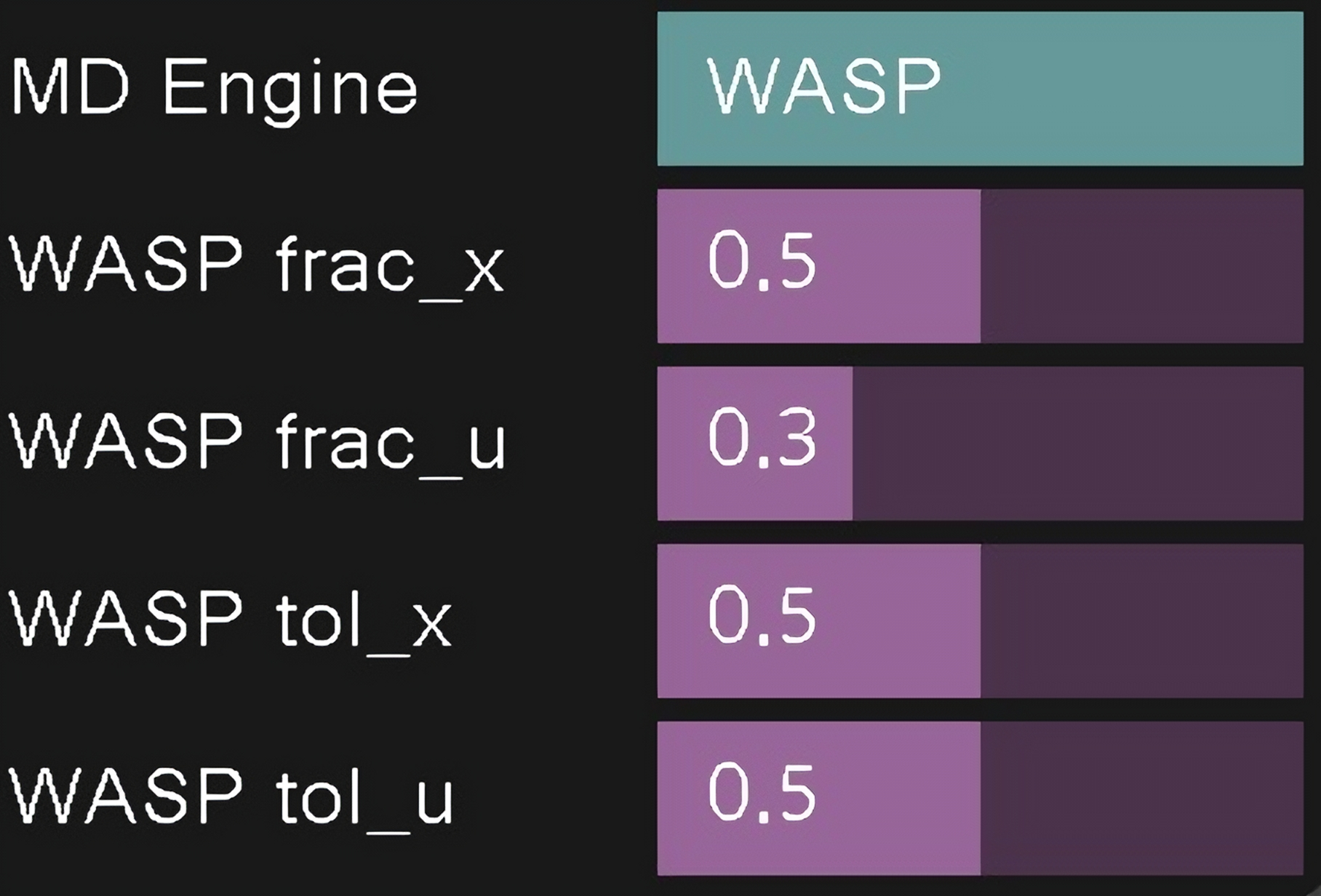}
    \caption{A screenshot of GUI components for users to activate WASP and control WASP's accuracy interactively}
    \label{fig:gui_bars}
    \vspace{-0.6cm}
\end{figure}

As shown in Figure~\ref{fig:gui_bars}, users can interactively adjust these two parameters for states ($\texttt{frac}_x$ and $\texttt{tol}_x$) and controls ($\texttt{frac}_u$ and $\texttt{tol}_u$) through slider bars in the MJPC GUI under ``Agent → Planner Settings''. The sliders range from 0 to 1: lower $\texttt{tol}$ and higher $\texttt{frac}$ values mean higher accuracy requirements and more iterations. As alluded to above, at maximum settings ($\texttt{frac}=1$, $\texttt{tol}=0$), WASP reduces to become equivalent to FD.

\section{Evaluation}
\label{sec:evaluation}

We demonstrate the efficacy of WASP derivatives in MJPC through three experiments.  In this section, we overview these experiments and present our results.  All experiments were executed on a desktop computer with an Intel i7 5.4GHz 28-core processor and 32 GB of RAM.




\subsection{Evaluation Tasks}
\label{subsec:task_selection}

We selected tasks from the MJPC benchmark suite based on three criteria: 
\begin{enumerate}
\item The task must satisfy $d_x + d_u \geq 10$ to ensure non-trivial computational complexity.
\item The task must be solvable using at least one derivative-based planner with an FD backend to establish a performance baseline.
\item The task must span a meaningful duration rather than being instantaneous to allow performance measurement over multiple planning iterations.
\end{enumerate}

Based on these criteria, we evaluated ten locomotion tasks spanning various robot morphologies and complexities as shown in Fig.~\ref{fig:teaser}:

\begin{itemize}
    \item \textbf{Quadrotor} ($d_x + d_u = 16$): A UAV achieving a set of positional goals while at the same time avoiding collisions with obstacles.
    \item \textbf{Swimmer} ($d_x + d_u = 26$): An underwater snake robot navigating through fluid dynamics.
    \item \textbf{Quadruped Tasks} ($d_x + d_u = 48$): Six distinct locomotion behaviors (stand, climb, walk, canter, trot, gallop) of a four-legged robot with complex contact dynamics.
    \item \textbf{Biped Balance} ($d_x + d_u = 48$): A four-legged robot dog maintaining upright posture with only its hind legs contact the ground.
    \item \textbf{Humanoid Walk} ($d_x + d_u = 75$): A humanoid robot executing stable walking gaits
\end{itemize}

These tasks represent diverse challenges in robotics control, from aerial navigation to legged locomotion with varying contact patterns. All tasks use default MJPC parameters unless otherwise specified, with planning horizon $T = 50$, timestep $\Delta t = 0.01$s, and the impedance ratio is set to $\text{impratio} = 100$, making the frictional constraints much stiffer relative to normal constraints to obtain high-fidelity contact dynamics.

\subsection{Experiment 1: WASP vs. FD}
\label{subsec:wasp_vs_fd}

In Experiment 1, we compare WASP and FD as derivative backends in MJPC.  Here, we outline the procedure, metrics, comparisons, and results for Experiment 1.

\subsubsection{Procedure}
\label{sec:procedure1}

For each task, we run the derivative-based planner(s) over a 30-second simulation window, recording performance metrics at each planning iteration.     

Parameters $\texttt{tol}_x$ and $\texttt{tol}_u$ were fixed at 0.5 in Experiment 1. Parameters $\texttt{frac}_x$ and $\texttt{frac}_u$ were minimally tuned ahead of time.  Specifically, we started with parameters $\texttt{frac}_x = \texttt{frac}_u = 0.3$ then incrementally raised both parameters until the robot first succeeds at the task.  Typically, task success is evident when the Performance Ratio (explained below) exceeds a value of $0.7$. 

\subsubsection{Metrics}
\label{sec:metrics1}
We evaluate performance using three primary metrics:
$$
\text{M.D. Speedup} = \frac{\text{avg. model derivative time with FD}}{\text{avg. model derivative time with WASP}}
$$

$$
\text{Speedup} = \frac{\text{avg. planning time with FD}}{\text{avg. planning time with WASP}}
$$
$$
\text{Performance Ratio} = \frac{\text{avg. cost with FD}}{\text{avg. cost with WASP}}
$$
where a speedup greater than 1 indicates WASP is faster, and a performance ratio close to 1 indicates comparable task performance.

\subsubsection{Comparisons}
We compare two planners in Experiment 1: (1) iLQG using an FD backend; and (2) iLQG using a WASP backend.  In preliminary experiments, we found that iLQG outperformed gradient descent on all of our tasks, so we only present these results here for simplicity.  

\subsubsection{Results}

\begin{table}[t!]
\centering
\caption{\small WASP vs. FD Performance Comparison}
\label{tab:wasp_vs_fd}
\adjustbox{width=\columnwidth}{
\begin{tabular}{l|c|c|c|c}
\toprule
\textbf{Task} & \textbf{WASP Params} & \textbf{M.D. Speedup} & \textbf{Speedup} & \textbf{Perf. Ratio} \\
&$(\texttt{frac}_x, \texttt{frac}_u)$  & & & \\
\midrule
Quadrotor  & (0.3, 0.3) & 1.92 & 1.17 & 1.02 \\
Swimmer  & (0.3, 0.3) & 1.53 & 1.13 &1.16 \\
Quadruped Stand &  (0.3, 0.3) & 2.08 & 1.45 & 1.22 \\
Quadruped Climb & (0.5, 0.5) & 1.61 & 1.31 &0.73 \\
Quadruped Walk  & (0.5, 0.5) & 1.57 &1.27& 0.70 \\
Quadruped Canter  & (0.4, 0.4) & 1.46 & 1.21 & 0.71 \\
Quadruped Trot& (0.5, 0.5) & 1.43& 1.21 & 0.75 \\
Quadruped Gallop & (0.4, 0.3) & 1.28 &1.12& 0.73 \\
Biped Balance & (0.8, 0.3) & 1.52 & 1.29&0.82 \\
Humanoid Walk  & (0.8, 0.6) & 1.26 & 1.10 & 0.90 \\
\bottomrule
\end{tabular}
}
\vspace{-0.7cm}
\end{table}

Results for Experiment 1 are shown in Table~\ref{tab:wasp_vs_fd}.  We see that WASP achieves speedups ranging from 1.26× to 2.08× compared to FD for model derivative computation across all tasks while maintaining at least sufficient task performance (defined here as Performance Ratio $\geq 0.7)$. Notably, for some tasks (Quadrotor, Swimmer, and Quadruped Stand), WASP achieves performance ratios greater than 1, indicating improved task performance despite using approximated derivatives.  We speculate that these results arise from slight approximation errors aiding in escaping local minima, though a deeper investigation is left for future work.


\subsection{Experiment 2: WASP-based iLQG vs. Sampling-Based Planners}
\label{subsec:wasp_vs_sampling}

In Experiment 2, we compare WASP-based iLQG with stochastic sampling-based planners in MJPC.  This section outlines the procedure, metrics, comparisons, and results for Experiment 2.


\subsubsection{Procedure}
The procedure in Experiment 2 is the same as Experiment 1 (\S\ref{sec:procedure1}) including all parameter settings.

\subsubsection{Metrics}
Our metrics in Experiment 2 are similar to those in Experiment 1 (\S\ref{sec:metrics1}): differing in two way: (1) we remove the M.D. Speedup metric as stochastic sampling-based planners do not take model derivatives; and (2) the Speedup metric replaces average planning time with FD in the numerator with average planning time for a given sampling-based planner.



A speedup greater than or close to 1 with performance ratio greater than 1 indicates the WASP-based iLQG is faster or as fast as sampling-based planners with better task performance.

\subsubsection{Comparisons}
We compare five planners in Experiment 2: (1) iLQG using a WASP backend; (2) Predicative Sampling~\cite{mjpc} (3) Robust Sampling~\cite{mppi}; (4) Cross Entropy Method~\cite{cem}; and (5) Sample Gradient~\cite{zoopt}.  Planners 2--5 are all of the stochastic sampling-based planners offered in MJPC.

\subsubsection{Results}
\begin{table*}[htbp]
\centering
\caption{\small WASP-Based iLQG vs. Sampling-Based Planners (A red ``X'' means the sampling-based planner fails to complete the task by exceeding a joint limit, falling over, or failing to reach the goal within $30$ seconds)}
\label{tab:wasp_vs_sampling}
\setlength{\tabcolsep}{8pt}
\begin{tabular}{l|c|cc|cc|cc|cc}
\toprule
\multirow{2}{*}{\textbf{Condition}} & \multirow{2}{*}{\textbf{\# Samples}} & \multicolumn{2}{c|}{\textbf{Predictive}} & \multicolumn{2}{c|}{\textbf{Robust}} & \multicolumn{2}{c|}{\textbf{Cross Entropy}} & \multicolumn{2}{c}{\textbf{Sample Gradient}} \\
\cmidrule{3-10}
 & & Speedup & Perf. & Speedup & Perf. & Speedup & Perf. & Speedup & Perf. \\
\midrule
Quadrotor & 30 & 1.32 & 1.36 & 2.39 & 1.40 & 1.35 & 1.93 & 1.18 & 1.08 \\
Swimmer & 32 & 1.38 & 1.58 & 1.90 & 2.03 & 1.31 & 1.65 & 1.26 & 1.24 \\ 
Quadruped Stand & 60 & 1.34 & 173.1\redcross & 2.49 & 442.5\redcross & 0.93 & 32.15 & 1.54 & 105.0\redcross \\
Quadruped Climb & 60 & 2.04 & 90.1\redcross & 3.19 & 115.6\redcross & 1.26 & 64.8\redcross & 1.84 & 54.2\redcross \\
Quadruped Walk & 60 & 1.78 & 265.3\redcross & 2.78 & 340.4\redcross & 1.10 & 190.8\redcross & 1.61 & 159.5\redcross \\
Quadruped Canter & 60 & 2.48 & 157.5\redcross & 3.87 & 202.1\redcross & 1.52 & 113.2\redcross & 2.24 & 94.7\redcross \\
Quadruped Trot & 60 & 2.15 & 106.1\redcross & 3.36 & 136.2\redcross & 1.33 & 76.3\redcross & 1.94 & 63.8\redcross \\
Quadruped Gallop & 60 & 2.59 & 113.8\redcross & 4.04 & 146.0\redcross & 1.59 & 81.8\redcross & 2.34 & 68.4\redcross \\
Biped Balance & 60 & 1.33 & 5.11\redcross & 2.20 & 4.86\redcross & 1.23 & 4.48\redcross & 1.01 & 4.82\redcross \\
Humanoid Walk & 100 & 0.85 & 3.29\redcross & 0.98 & 3.83\redcross & 0.85 & 8.85\redcross & 0.85 & 2.25\redcross \\
\bottomrule
\end{tabular}
\vspace{-0.4cm}
\end{table*}
Results for Experiment 1 are shown in Table~\ref{tab:wasp_vs_sampling}.  Here, WASP-based iLQG significantly outperforms sampling-based planners on these contact-rich tasks. A red cross next to the performance ratio indicates task failure by the sampling-based planner. For quadruped tasks, biped balance and humanoid walk, iLQG with WASP achieves up to 4.0× speedups while sampling-based planners struggles to complete the tasks, demonstrating that gradient-based optimization with WASP is particularly effective for locomotion tasks with complex contact dynamics. 

\subsection{Experiment 3: Parameter Robustness Analysis}
\label{subsec:parameter_robustness}

In Experiment 3, we assess WASP's sensitivity to parameter selection.  This section outlines the procedure, metrics, comparisons, and results.  

\subsubsection{Procedure}
We only assess the quadruped trotting task over 1000 planning iterations, recording performance metrics at each iteration.


\subsubsection{Metrics}
Our metrics in Experiment 3 are: (1) \textit{Cost evolution}: Task performance as measured by cost function over iterations; (2) \textit{Computation time}: Model derivative calculation time in milliseconds; and (3) \textit{Simulation steps}: Number of forward simulations required.

\subsubsection{Comparisons}
We compare five parameter configurations against FD baseline:
\begin{itemize}
    \item $(\texttt{frac}_x, \texttt{frac}_u) = (0.5, 0.5)$: Balanced baseline.
    \item $(\texttt{frac}_x, \texttt{frac}_u) = (0.5, 0.3)$: Reduced action accuracy.
    \item $(\texttt{frac}_x, \texttt{frac}_u) = (0.5, 0.1)$: Minimal action accuracy.
    \item $(\texttt{frac}_x, \texttt{frac}_u) = (0.3, 0.5)$: Reduced state accuracy.
    \item $(\texttt{frac}_x, \texttt{frac}_u) = (0.1, 0.5)$: Minimal state accuracy.
\end{itemize}

\subsubsection{Results}

\begin{figure*}[t!]
\centering
\includegraphics[width=0.9\textwidth]{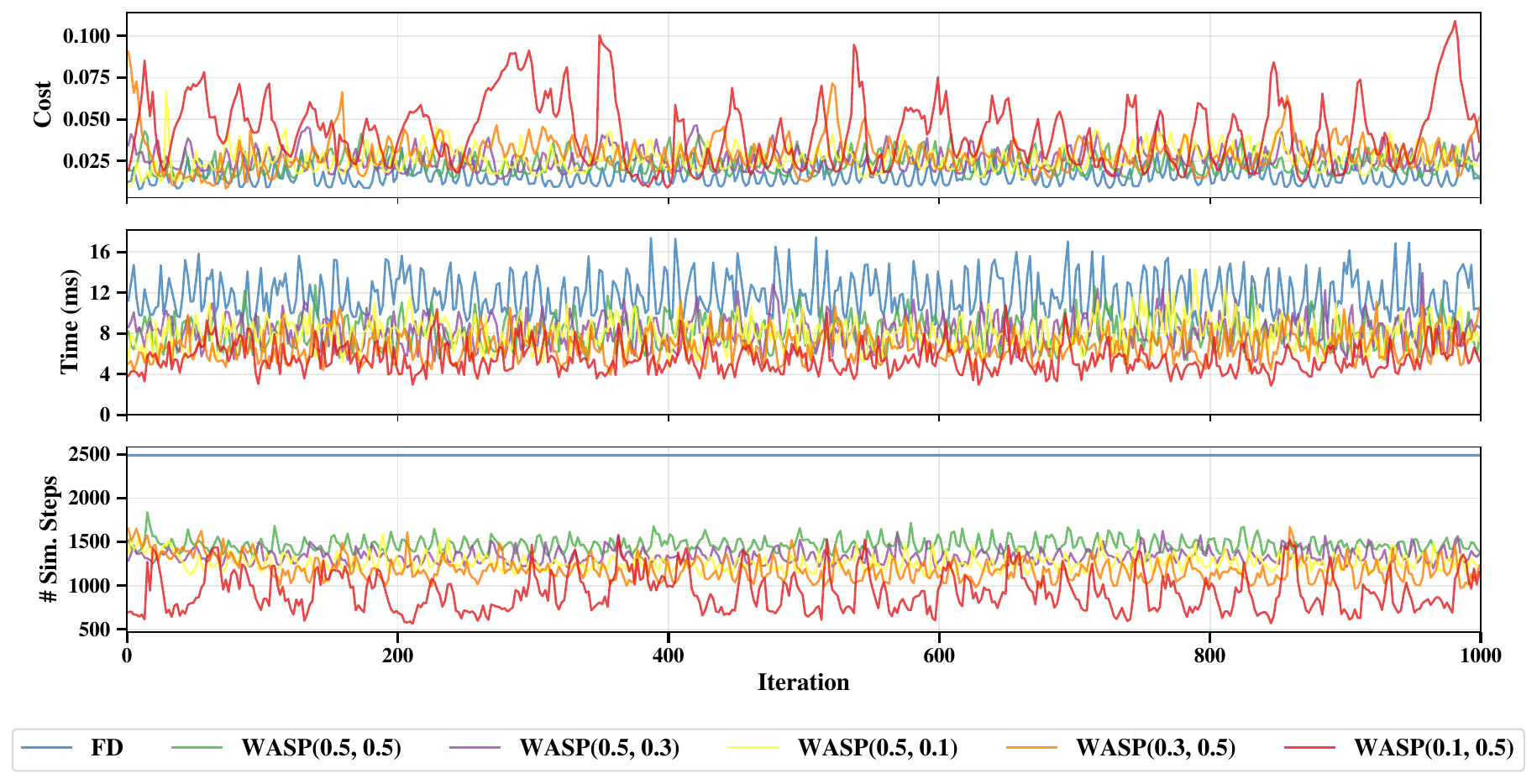}
\caption{Parameter sensitivity analysis of WASP on the quadruped trotting task. 
(a) Cost convergence across settings, with performance degrading only under very low state accuracy. 
(b) Average computation time per derivative evaluation, consistently lower than FD. 
(c) Number of forward simulations, showing adaptive resource usage with accuracy. 
Overall, WASP maintains stable performance and computational savings across a broad parameter range, with state accuracy more critical than control accuracy.}
\label{fig:parameter_robustness}
\vspace{-0.6cm}
\end{figure*}


Results for Experiment 3 are shown in Figure~\ref{fig:parameter_robustness}.  At a high level, we see that WASP is reasonably robust to parameter variations. When reducing $\texttt{frac}_u$ while maintaining $\texttt{frac}_x = 0.5$ (purple and yellow lines), the planner maintains stable performance with consistent computational savings. However, reducing $\texttt{frac}_x$ while maintaining $\texttt{frac}_u = 0.5$ (orange and red lines) causes instability, evidenced by cost spikes and erratic simulation counts. These results suggest that iLQG is particularly sensitive to the state transition accuracy, consistent with prior analysis by Todorov et al.~\cite{ilqg_error}. 


\section{Discussion}
\label{sec:discussion}

Our work demonstrates that replacing finite differencing (FD) with Web of Affine Spaces (WASP) derivatives substantially improves the efficiency of derivative-based planning in MJPC. Across a broad set of locomotion tasks, WASP consistently reduced the computational burden of model derivative evaluations while maintaining, and in some cases even improving, task performance. The resulting speedups for model derivative computation of 1.26--2.08$\mathsf{x}$ enabled iLQG-based planners to operate faster than stochastic sampling-based planners, while delivering stronger and more reliable control. These findings suggest that coherence-based derivative approximations can offer a compelling balance between efficiency and robustness in iterative control settings.

Beyond efficiency, our evaluation highlights how approximate derivatives can at times improve task outcomes compared to exact FD. We speculate that mild approximation error introduces a form of regularization, smoothing sharp gradients that often arise in contact-rich environments and helping optimization escape poor local minima. This effect, while encouraging, warrants deeper study in future work to fully understand its implications. In addition, our parameter sensitivity analysis showed that WASP is generally robust to variations, with accuracy in state transitions being more critical than control accuracy, a finding that aligns with prior analyses of iLQG.

Importantly, WASP integrates directly into MJPC as a drop-in replacement, requiring no changes to the simulator's source code and allowing users to seamlessly switch between FD and WASP backends. This design lowers the barrier to adoption and creates opportunities for broader use in research and practice, particularly in scenarios where rapid derivative evaluations are essential for real-time performance. To support adoption, we also provide a fully open-source implementation of MJPC with WASP derivatives, making it straightforward for practitioners and researchers to experiment with these capabilities in their own applications.

\subsection{Limitations and Future Directions}

While promising, this study has several limitations. First, all experiments were conducted in simulation, primarily on locomotion benchmarks. Since our contribution concerns derivative computation within a simulator-based MPC stack, simulation provides the appropriate controlled setting for measuring computational cost and numerical stability without hardware confounds. We therefore treat sim-to-real transfer as orthogonal to the present work, though extending to physical systems is a natural next step.

Second, both FD and WASP-based gradient planners struggled on contact-rich manipulation tasks, where short MPC horizons and discontinuous contact dynamics often lead to failures. These difficulties appear largely independent of the derivative approximation, pointing to structural limitations of short-horizon gradient-based MPC. Addressing them will likely require architectural changes, such as longer horizons or hybrid derivative- and sampling-based methods.

Finally, WASP's accuracy parameters require manual tuning to balance speed and fidelity. Developing principled or adaptive selection schemes could further improve robustness and usability.

Overall, we view this work as a step toward integrating coherence-based derivative approximations into real-time MPC. By demonstrating meaningful speed and stability gains within MJPC, we aim to expand the practical alternatives to finite differencing and encourage broader exploration of structured approximate derivatives in robotics control.

\bibliographystyle{plainnat}

\bibliography{refs}


\end{document}